\documentclass[sigconf]{acmart}
\AtBeginDocument{%
  }

\copyrightyear{2026}
\acmYear{2026}
\setcopyright{cc}
\setcctype{by}
\acmConference[WSDM '26]{Proceedings of the Nineteenth ACM International Conference on Web Search and Data Mining}{February 22--26, 2026}{Boise, ID, USA}
\acmBooktitle{Proceedings of the Nineteenth ACM International Conference on Web Search and Data Mining (WSDM '26), February 22--26, 2026, Boise, ID, USA}
\acmPrice{}
\acmDOI{10.1145/3773966.3779401}
\acmISBN{979-8-4007-2292-9/2026/02}

\usepackage{xcolor}
\usepackage{balance}
\usepackage{mathtools}
\usepackage{url}
\usepackage{amsfonts}
\usepackage{algorithmic}
\usepackage{textcomp}
\usepackage{algorithm}
\usepackage{multirow}
\usepackage{color}
\usepackage{subcaption}
\usepackage{enumitem}

\usepackage{tikz}
\usepackage{pgfplots,pgfplotstable}

\usepackage{pdfrender}

\pgfplotsset{compat=1.5.1}
\newenvironment{customlegend}[1][]{%
  \begingroup
  \csname pgfplots@init@cleared@structures\endcsname
  \pgfplotsset{#1}%
}{%
  \csname pgfplots@createlegend\endcsname
  \endgroup
}%
\def\addlegendimage{\csname pgfplots@addlegendimage\endcsname}
\usetikzlibrary{patterns}
\usetikzlibrary{pgfplots.groupplots}
\usetikzlibrary{
  pgfplots.colorbrewer,
}
\pgfplotsset{
  cycle list/.define={my marks}{
    every mark/.append style={solid,fill=\pgfkeysvalueof{/pgfplots/mark list fill}},mark=*\\
    every mark/.append style={solid,fill=\pgfkeysvalueof{/pgfplots/mark list fill}},mark=square*\\
    every mark/.append style={solid,fill=\pgfkeysvalueof{/pgfplots/mark list fill}},mark=triangle*\\
    every mark/.append style={solid,fill=\pgfkeysvalueof{/pgfplots/mark list fill}},mark=diamond*\\
  },
}
\definecolor{aa}{rgb}{0.2,0.7,0.310}
\definecolor{cc}{rgb}{0.914,0.725,0.431}
\definecolor{bb}{rgb}{0.514,0.325,0.831}

\definecolor{xxx}{rgb}{0.6,0.80,0.98}
\definecolor{oxx}{rgb}{0.26,0.52,0.80}
\definecolor{oox}{rgb}{0.01,0.53,0.82}
\definecolor{ooo}{rgb}{0.00,0.23,0.6}

\definecolor{ooo_2}{HTML}{9b2948}
\definecolor{oox_2}{HTML}{ff7251}
\definecolor{oxx_2}{HTML}{ffca7b}
\definecolor{xxx_2}{HTML}{ffedbf}

\definecolor{comp}{HTML}{45C5F4}
\definecolor{la}{HTML}{92D050}
\definecolor{send}{HTML}{92D050}
\definecolor{wait}{HTML}{FFCD33}
\definecolor{recv}{HTML}{8D59B3}

\definecolor{s8}{HTML}{FF5349}
\definecolor{s16}{HTML}{FFCD33}
\definecolor{s32}{HTML}{92D050}
\definecolor{s64}{HTML}{45C5F4}
\definecolor{s128}{HTML}{8D59B3}

\definecolor{rotate}{HTML}{073b4c}
\definecolor{complex}{HTML}{6262FE}
\definecolor{house}{HTML}{8EC1DB}
\definecolor{tucker}{HTML}{118ab2}
\definecolor{plogicnet}{HTML}{F78C6B}
\definecolor{rnnlogic}{HTML}{ef476f}
\definecolor{compgcn}{HTML}{80B563}

\newcommand{\method}{\textsf{PROBE}}
\newcommand{\codeurl}{\url{https://github.com/mindslab-cau/probe-wsdm26}}

\begin{document}


\title[How Sharp and Bias-Robust is a Model? Dual Evaluation Perspectives on Knowledge Graph Completion]{How Sharp and Bias-Robust is a Model? \\Dual Evaluation Perspectives on Knowledge Graph Completion}

% Revisiting Evaluation of Knowledge Graph Completion}

\author{Sooho Moon}
\affiliation{%
  \institution{Chung-Ang University}
  \city{Seoul}
  \country{Korea}}
\email{moonwalk725@cau.ac.kr}

\author{Yunyong Ko}
\authornote{Corresponding author}% \authornotemark[1]
\affiliation{%
  \institution{Chung-Ang University}
  \city{Seoul}
  \country{Korea}}
\email{yyko@cau.ac.kr}
\orcid{0000-0003-1283-4697}


\begin{abstract}
Knowledge graph completion (KGC) aims to predict missing facts from the observed KG.
While a number of KGC models have been studied, the evaluation of KGC still remain underexplored. 
In this paper, we observe that existing metrics overlook two key perspectives for KGC evaluation:
(\textbf{A1}) \textit{predictive sharpness} -- the degree of strictness in evaluating an individual prediction, 
and (\textbf{A2}) \textit{popularity-bias robustness} -- the ability to predict low-popularity entities.
Toward reflecting both perspectives, 
we propose a novel evaluation framework (\textbf{{\method}}),
which consists of a rank transformer (RT) estimating the score of each prediction based on a required level of predictive sharpness and a rank aggregator (RA) aggregating all the scores in a popularity-aware manner.
Experiments on real-world KGs reveal that existing metrics tend to over- or under-estimate the accuracy of KGC models, whereas {\method} yields a comprehensive understanding of KGC models and reliable evaluation results.

% For reproducibility, we have released the code and datasets at {\codeurl}.
\end{abstract}

\begin{CCSXML}
<ccs2012>
   <concept>
       <concept_id>10002951.10003227.10003351</concept_id>
       <concept_desc>Information systems~Data mining</concept_desc>
       <concept_significance>500</concept_significance>
       </concept>
   <concept>
       <concept_id>10010147.10010178.10010187</concept_id>
       <concept_desc>Computing methodologies~Knowledge representation and reasoning</concept_desc>
       <concept_significance>500</concept_significance>
       </concept>
 </ccs2012>
\end{CCSXML}

\ccsdesc[500]{Information systems~Data mining}
\ccsdesc[500]{Computing methodologies~Knowledge representation and reasoning}

\keywords{Knowledge graph completion, Rank-based evaluation}

\maketitle

\section{Introduction}\label{sec-intro} % 2
A knowledge graph (KG) is a graph-structured real-world knowledge,
where each knowledge fact has the form of a triple $(h,r,t)$.
KGs~\cite{bollacker2008freebase,suchanek2007yago,miller1995wordnet,carlson2010toward} have been widely used in a range of applications such as question answering (QA)~\cite{hao2017end,yih2014semantic}, news classification~\cite{ko2023khan,wang2018dkn},
drug discovery~\cite{zhang2021drug,zeng2022toward,bonner2022review}, and retrieval-augmented generation (RAG)~\cite{edge2024local,guo2020survey,pan2024unifying}. 
Real-world KGs, however, are inherently incomplete~\cite{trouillon2016complex,sun2019rotate}, 
i.e., a number of facts are missing, which can hinder the potential of KGs. 
To this limitation, \textit{knowledge graph completion} (KGC) has been widely studied~\cite{bordes2013translating,yang2015embeddingentitiesrelationslearning,trouillon2016complex,sun2019rotate,qu2020rnnlogic,qu2019probabilistic},
which aims to infer missing facts based on the observed KG structure, i.e., link prediction on KGs.

\begin{figure}[t]
\centering
\begin{tabular}{cc}
    \includegraphics[width=0.465\linewidth]{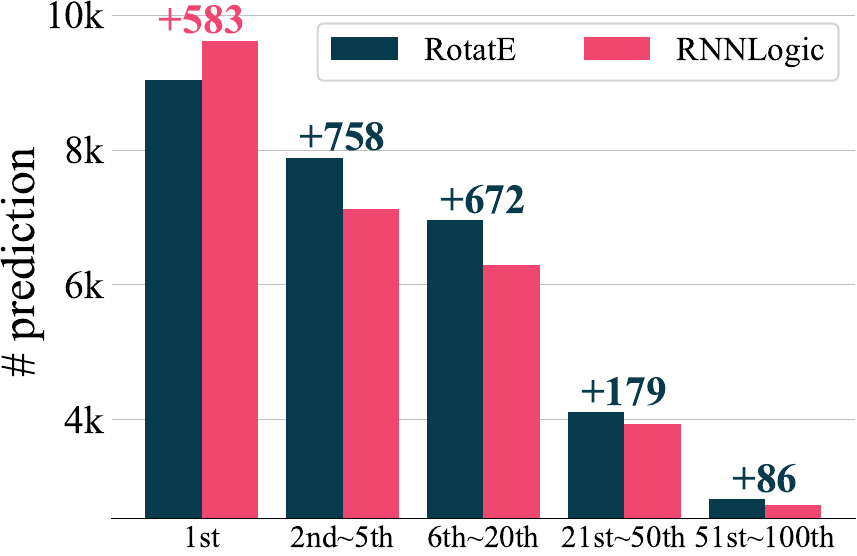} &
    \includegraphics[width=0.465\linewidth]{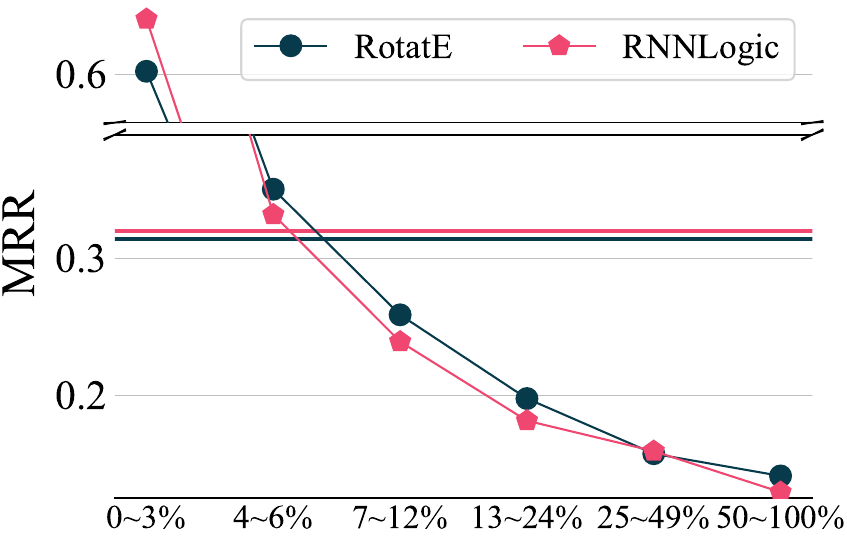} \\
    (a) Observation 1 & (b) Observation 2
\end{tabular}
\vspace{-3mm}
\caption{Observations: Existing rank-based metrics have overlooked two subtle yet critical perspectives (P1) and (P2).}
\vspace{-5mm}
\label{fig:aspect1}
\end{figure}

Despite recent breakthroughs in KGC models, little attention has been paid to the \textit{evaluation} of KGC.
To evaluate KGC models, \textit{rank-based} evaluation metrics (e.g., mean reciprocal rank (MRR) and Hits@K) are commonly adopted~\cite{hoyt2022unified,yang2022rethinking,mohamed2020popularity} due to the \textit{open-world assumption} (OWA)~\cite{yang2022rethinking,hoyt2022unified},
where the absence of a triple in a KG does not necessarily imply that it is false.
Existing metrics, however, overlook two subtle but critical perspectives in KGC evaluation.
% \textbf{(A1)} \textit{predictive sharpness} and \textbf{(A2)} \textit{popularity-bias robustness}.

\vspace{1mm}
\noindent
\textbf{(P1)} \textbf{Predictive sharpness.} 
We may answer the question “\textit{how strictly should we evaluate each individual prediction?}" differently depending on circumstances. 
For example, when we leverage biomedical KGs for drug discovery, 
less faithful facts may result in serious risks (e.g., harmful side effects or substantial costs for clinical trials~\cite{zeng2022toward}). 
Thus, in this case, a high level of \textit{predictive sharpness} is required in evaluating KGC models.
In other words, as a prediction becomes less accurate, a larger penalty should be imposed.
While, when commonsense KGs are used in recommender systems, 
a relatively low level of predictive sharpness may be acceptable as long as the predicted facts are beneficial for user and item modeling.

Such a perspective, however, has not been considered in existing metrics.
Figure~\ref{fig:aspect1}(a) shows the number of predictions of two state-of-the-art KGC models (RotatE~\cite{sun2019rotate} and RNNLogic~\cite{qu2020rnnlogic}) across different ranks.
RNNLogic outperforms RotatE in the 1st rank, while RotatE does RNNLogic in others (from 2nd to 100th ranks),
which implies that RNNLogic (resp. RotatE) is preferred when a high (resp. low) level of predictive sharpness is required.
However, RNNLogic alwasys gets a higher score than RotatE in existing metrics.

\vspace{1mm}
\noindent
\textbf{(P2)} \textbf{Popularity-bias robustness.} 
Real-world graphs generally follow a power-law degree distribution~\cite{wang2019tackling,xiong2018oneshot,ko2021mascot,jang2023sage}, i.e., in a KG, most entities appear in only a few triples (i.e., low popularity), while a small number of entities appear in a large number of triples (i.e., high popularity). 
Thus, a KGC model may exhibit \textit{popularity bias} since it trains high-popularity entities much more frequently than low-popularity ones.
To verify this hypothesis, we measure the KGC accuracy (MRR) of RotatE and RNNLogic with respect to popularity.
As shown in Figure~\ref{fig:aspect1}(b), 
both models exhibit strong popularity bias (i.e., much higher accuracy on high-popularity entities than on low-popularity ones).
Interestingly, although RotatE outperforms RNNLogic in all groups except for the highest-popularity group (i.e., RotatE is more robust to popularity bias than RNNLogic), 
RNNLogic gets a higher score,
which implies that existing metrics do not consider the perspective of popularity bias

To reflect both perspectives (P1) and (P2), we propose a novel framework for KGC evaluation, 
named \textbf{\underline{P}}redictive sha\textbf{\underline{R}}pness and p\textbf{\underline{O}}pularity-\textbf{\underline{B}}ias robustness aware \textbf{\underline{E}}valuation (\textbf{{\method}}).
For every test triple, {\method} (1) transforms the predicted rank of the missing entity into a score, based on varying levels of predictive sharpness, 
and then (2) computes the final score by taking popularity-aware weighted average of all the transformed scores.

Extensive experiments on real-world KGs reveal that existing rank-based metrics tend to:
(1) over-estimate KGC models that make a small number of perfect predictions, i.e., 1st rank,
while under-estimate those that consistently make not perfect but good predictions, i.e., 2nd-5th ranks,
and (2) over-estimate KGC models with strong popularity bias, which often perform poorly on low-popularity entities.
These results indicate that existing rank-based metrics have limitations in evaluating KGC models --i.e., overlooking the critical perspectives (P1) and (P2).
While {\method} evaluates KGC models from the perspectives of required levels of predictive sharpness and popularity-bias robustness.
Moreover, we provide guidelines for determining appropriate levels of predictive sharpness and popularity-bias robustness, depending on circumstances.

% To the best of our knowledge, this is the first work to incorporate the two key perspectives (P1) and (P2) into the KGC evaluation.
\noindent
\textbf{Contributions}. The main contributions of this work are as follows.
\begin{itemize}[leftmargin=10pt]
    \item \textbf{Observations}: We observe that existing KGC evaluation metrics have overlooked subtle yet critical perspectives: (P1) predictive sharpness and (P2) robustness to popularity-bias.
    \item \textbf{Framework}: We propose a novel evaluation framework, {\method}, that evaluates KGC models by considering the varying levels of (P1) predictive sharpness and (P2) popularity-bias robustness.
    \item \textbf{Evaluation}: Via extensive experiments on two real-world KGs, 
    we demonstrate that {\method} provides a comprehensive understanding of KGC models and reliable evaluation results.

\end{itemize}

\noindent
For reproducibility, we have released the code and datasets at {\codeurl}.

\section{Problem Definition}\label{sec-proposed-preliminary}
% First, we introduce the notations and define the problem that we consider. 
% The notations used in this paper are described in Table~\ref{table:notations}.
We consider the following problem, knowledge graph completion.

\vspace{1mm}
\noindent
\textbf{\textsc{Problem 1}} (\textsc{Knowledge Graph Completion}).
Given a knowledge graph (KG) $\mathcal{G}=(\mathcal{E},\mathcal{R},\mathcal{T})$,
where $\mathcal{E}$ is the set of entities, $\mathcal{R}$ is the set of relations, and $\mathcal{T}=\{((h,r,t)| h,t \in \mathcal{E}, r \in \mathcal{R}\}$ is the set of triples (i.e., facts),
the goal of knowledge graph completion (KGC) is to infer missing facts based on the observed KG.

Given a KGC model $\theta(h,r,t)$, 
its parameters are trained so that positive triples obtain higher scores than negative triples.

\section{Proposed Framework: {\method}}\label{sec-proposed} % 5.5 ~ 6
% In this section, we present a generalized framework for KGC evaluation, 
% named \textbf{{\method}}, which stands for \textbf{\underline{P}}redictive sha\textbf{\underline{R}}pness and p\textbf{\underline{O}}pularity-\textbf{\underline{B}}ias robustness aware \textbf{\underline{E}}valuation (\textbf{{\method}}).
% First, we introduce the notations and define the problem that we consider (Section~\ref{sec-proposed-preliminary}). 
% Then, we describe two key components of {\method}: 
% a rank transformer (Section~\ref{sec-proposed-rt}) and a rank aggregator (Section~\ref{sec-proposed-ra})
% Finally, we theoretically analyze the properties of {\method} (Section~\ref{sec-proposed-theoritical})

\noindent
\textbf{\underline{Overview of {\method}}}.
{\method} evaluates KGC models by using the same evaluation protocol used in~\cite{sun2020reevaluation,qu2020rnnlogic,li2022house,bordes2013translating}.
Given a trained model $\theta(\cdot)$  and a set of test triples $\mathcal{T}_{test}$, 
{\method} consists of the three steps: (1) prediction, (2) transformation, and (3) aggregation.

\vspace{1mm}
\noindent
\textbf{(1) Prediction}: 
For each test triple $(h,r,t)  \in \mathcal{T}_{test}$, two queries are generated by masking either the head or tail entity, i.e., $(h,r,?) $ and $(?,r,t)$.
Then, a KGC model $\theta(\cdot)$ estimates the probability of being the missing entity for all possible candidate entity $e' \in \mathcal{E}$
The candidate entities are sorted in descending order, based on their scores,
and the rank of the actual missing entity is determined, i.e., $1\leq r\leq|\mathcal{E}|$.
For all test triples, a set of ranks $\mathbf{r}=\{r_1,r_2,...,r_{|\mathcal{T}_{{test}}|*2}\}$ is produced ($ \because$ two queries for each test triple).

\vspace{0.5mm}
\noindent
\textbf{(2) Transformation}: 
Then, given a set of ranks $\mathbf{r}$
A rank transformation function $f(r):(\mathbb{N}^1  \rightarrow \mathbb{R}^1)$ converts each rank $r \in \mathbf{r}$ into a score.
Typically, the function $f(\cdot)$ is defined to be \textit{anti-monotone}, so that lower rank values (i.e., correct predictions) yield higher scores, whereas higher rank values yield lower scores (e.g., MRR: $f(r)=1/r$).
The step produces a new set of scores $\mathbf{c}=\{c_1,c_2,...,c_{|\mathcal{T}_{test}|*2}\}$.

\vspace{0.5mm}
\noindent
\textbf{(3) Aggregation}: 
Finally, given a set of scores $\mathbf{c}$,
a rank aggregator $agg: (\mathbb{R}^{|\mathcal{T}_{test}|*2} \rightarrow \mathbb{R}^{1})$ computes the overall KGC accuracy by taking the (weighted) average of all scores in $\mathbf{c}$ (e.g., MRR assigns the same weight to all scores).
% strat-MRR~\cite{mohamed2020popularity}
% different weights based on popularity (e.g., ).

% Based on this process, 
% {\method} aims to comprehensively evaluate KGC models in the aspects of varying levels of (A1) predictive sharpness and (A2) popularity-bias robustness.
% We will describe a \textbf{rank transformer} (\textbf{RT}) and a \textbf{rank aggregator} (\textbf{RA}) of {\method} in Sections~\ref{sec-proposed-rt} and~\ref{sec-proposed-ra}.

\begin{figure}[t]
\centering
\begin{tabular}{cc}
    \includegraphics[width=0.44\linewidth]{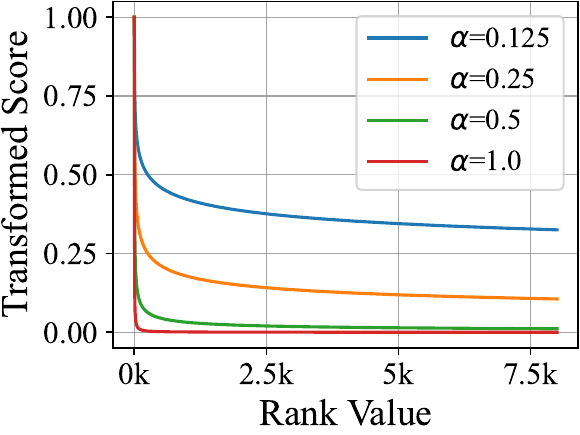} &
    \includegraphics[width=0.44\linewidth]{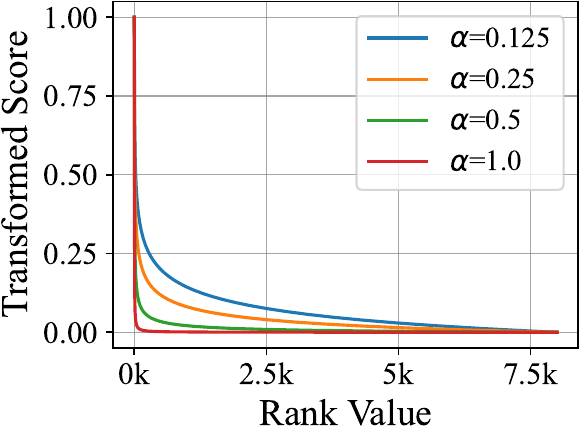} \\
    (a) Original RT  & (b) Affine RT
\end{tabular}
\vspace{-4mm}
\caption{Rank transformers of {\method}: Affine RT ensures that all scores are in the range [0,1].}
\vspace{-6mm}
\label{fig:RT}
\end{figure}

% \subsection{Rank Transformer (RT)}\label{sec-proposed-rt}
% In this section, we describe the rank transformer (RT) of {\method}, 
% which can reflects varying levels of (P1) predictive sharpness.

\vspace{0.5mm}
\noindent
\textbf{\underline{Rank transformation}.}
The RT of {\method} converts the rank of each prediction into a score, 
considering the required level of predictive sharpness. 
Specifically, it takes the predicted rank $r \in \mathbf{r}$ and the \textit{sharpness control factor} $\alpha$ and computes its corresponding score:
\begin{equation}
    f(r,\alpha)={1 \over r^{\alpha}} \hspace{1mm} (r \in\mathbf{r})\label{eq:transform}.
\end{equation}
% where $\mathbf{r}$ is a set of ranks predicted by a KGC model that {\method} aims to evaluate and 
% where $\alpha$ is a hyperparameter to control the level of predictive sharpness.
We note that this formulation can generalize diverse rank transformation rules of existing metrics.
For example, if $\alpha=1$, it is the same as the rank transformation function of MRR (i.e., the reciprocal of a rank),
while if $\alpha=-1$, it degenerates to the rank transformation function of MR (i.e., using the rank itself).
Thus, existing metrics can be viewed as special cases with a \textit{fixed} level of predictive sharpness.

When $\alpha<0$, the RT is \textit{sensitive} to outliers (e.g., large rank values of wrong predictions), 
which leads to incorrect evaluation~\cite{hoyt2022unified}.
That's the reason why a transformation function is commonly defined as an anti-monotone function.
With the same reason, we consider only when $\alpha>0$.
Figure~\ref{fig:RT}(a) shows the transformed scores of the RT according to the sharpness factor values.
If $\alpha$ gets larger, i.e., higher level of the predictive sharpness, incorrect predictions are heavily penalized.
Otherwise, i.e., lower level of the predictive sharpness, incorrect predictions are less penalized.

\vspace{0.5mm}
\noindent
\textbf{\underline{Distinguishability of RT}}.
As shown in Figure~\ref{fig:RT}(a), the lower bound of a RT increases as $\alpha$ approaches zero.
This causes the score range to shrink, i.e., ${1 \over |\mathcal{E}|^{\alpha} }<f(r,\alpha)<1$, where $|\mathcal{E}|$ is the number of entities,
which hinders distinguishing relative differences among KGC models and thereby reduces its effectiveness as an evaluation metric.
This issue also occurs when $|\mathcal{E}|$ is small.

To address this issue, we apply an \textit{affine} transformation to our RT function $f(\cdot)$,
which ensures that all scores are rescaled to the range $[0,1]$, rather than $[{1 \over |\mathcal{E}|^{\alpha} },1]$ (See Figure~\ref{fig:RT}(b)).
The final rank transformer of {\method} is defined as:
\begin{equation}
    f^*(r,\alpha)=\frac{f(r,\alpha)-1}{1-(|\mathcal{E}|^{-\alpha})}+1\label{eq:transform-affine}.
\end{equation}
This form in Eq.~\ref{eq:transform-affine} enhances the distinguishability of {\method}, maintaining relative ranks,
and thereby satisfies key properties required for serving as a rank-based evaluation metric for KGC models.

\vspace{1mm}
\noindent
\textbf{\textsc{Property 1}} (\textsc{Fixed optimum}).
{\method} satisfies the \textit{fixed optimum} property so that it assigns the optimum score $c_{opt}$ when a model makes a correct prediction (i.e., 1st rank, $f^*(1)=c_{opt}=1$).

\vspace{1mm}
\noindent
\textbf{\textsc{Property 2}} (\textsc{Fixed pessimum}).
{\method} satisfies the \textit{fixed pessimum} property so that it assigns the pessimum score $c_{pes}$ when a model makes the worst prediction (i.e., $|\mathcal{E}|$th rank, $f^*(|\mathcal{E}|)=c_{pes}=0$).

\vspace{1mm}
We note that existing metrics (e.g., MRR, p-MRR and Strat-MRR) do not satisfy the \textsc{Property 2},
instead, they only provide \textit{asymptotic} lower bounds, which are less practical than fixed lower bounds.
This highlights the superior distinguishability of {\method} over existing metrics, especially for incorrect predictions.

\begin{figure}[t]
\centering
\begin{tabular}{cc}
    \includegraphics[width=0.44\linewidth]{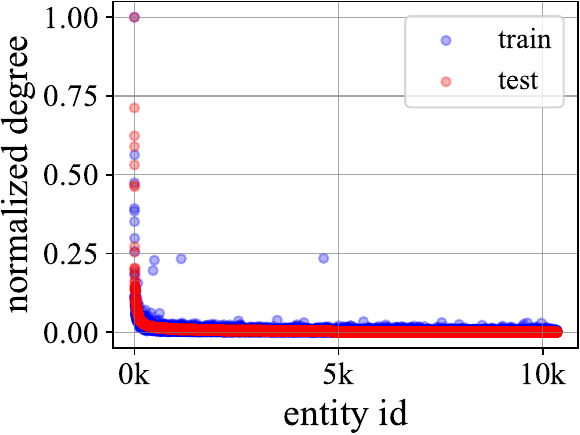} &
    \includegraphics[width=0.44\linewidth]{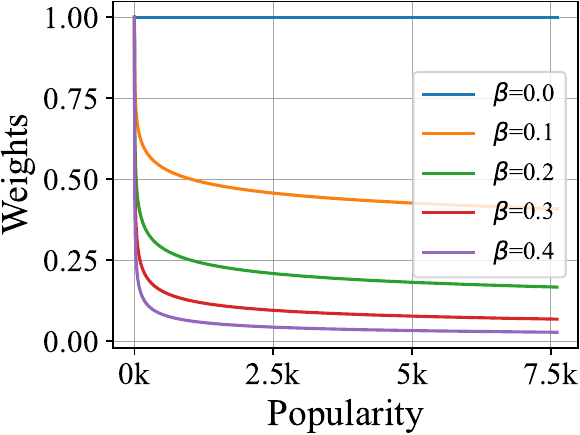} \\
    (a) Entity degree distributions & (b) Weight functions
\end{tabular}
\vspace{-3mm}
\caption{(a) Similar distributions in training and test sets and (b) weight functions with varying popularity-bias robustness.}
\vspace{-5mm}
\label{fig:RA}
\end{figure}

% \subsection{Rank Aggregator (RA)}\label{sec-proposed-ra}
% In this section, we describe the rank aggregator (RA) of {\method}, 
% which can reflects varying levels of (P2) popularity-bias robustness.

\vspace{1mm}
\noindent
\textbf{\underline{Rank aggregation}.}
The RA of {\method} assigns a weight to each prediction based on the popularity of the missing entity to predict.
Intuitively, we assign a lower weight to a prediction if its missing entity has higher popularity in the training data, 
as high-popularity entities are likely to be sufficiently learned during training, which can lead to popularity bias in evaluation.
Figure~\ref{fig:RA}(a) shows that entity popularity (i.e., degree) distributions in the training and test sets are very similar, 
which further supports the validity of our weighting strategy.
Formally, the weight of the prediction for a test query $q=(h,r,?)$ or $(?,r,t)$ is defined as:
\begin{align}
    w_q = {1 \over (\epsilon + \delta(q)_{train})^{\beta}}\label{eq:weight},
\end{align}
where $\delta(q)_{train}$ is the popularity of the query $q=(h,r,?)$ or $(?,r,t)$, i.e., the number of training triples in which the missing entity appears,
$\beta$ is a factor to control the level of popularity-bias robustness,
and $\epsilon$ is a small constant to prevent from division by zero.

Figure~\ref{fig:RA}(b) shows the weight function according to the varying levels of popularity-bias robustness $\beta$,
where a larger $\beta$ penalizes the weights of high-popularity entities more severely.
Note that given the same triple $(h,r,t)$, 
there are two queries $(h,r,?)$ and $(?,r,t)$, and they can be assigned as different weights since the head and tail entities can be different.
Finally, all scores $\mathbf{c}$ transformed in the RT are averaged with their weights $\mathbf{w}$ to compute the accuracy of a KGC model, which is defined as:
\begin{align}
    {1 \over W}\sum^{|\mathcal{T}_{test}|*2}_{i=1} w_i \cdot c_i,\label{eq:aggregation}.
\end{align}

\section{Experimental Validation}\label{sec-eval} 
In this section, we aim to answer the following research questions: 
\begin{itemize}[leftmargin=8pt]
    \item \textbf{RQ1}. To what extent does (P1) the predictive sharpness affect the accuracy of KGC models?
    \item \textbf{RQ2}. To what extent  does (P2) the popularity-bias robustness affect the accuracy of KGC models?
    \item \textbf{RQ3}. Does {\method} enable a more comprehensive evaluation of KGC models than existing metrics?
\end{itemize}

\subsection{Experimental Setup}\label{sec:eval-setup}

\vspace{1mm}
\noindent
\textbf{KG datasets and KGC models.}
In our experiments, we use two real-world knowledge graphs (KGs), which were also used in~\cite{sun2019rotate,qu2020rnnlogic,trouillon2016complex,balazevic2019tucker,li2022house,dettmers2018convolutional,yang2015embeddingentitiesrelationslearning}:
FB15k237~\cite{toutanova2015observed} and (2) WN18RR~\cite{miller1995wordnet}.
Table~\ref{table:datasets} shows the data statistics.
We use four state-of-the-art KGC models:
2 embedding-based models (RotatE~\cite{sun2019rotate} and TuckEr~\cite{balazevic2019tucker}) and 2 rule-based models (pLogicNet~\cite{qu2019probabilistic} and RNNLogic~\cite{qu2020rnnlogic}).
For all KGC models, we use the official source codes provided by the authors\footnote{We release the code and datasets at {\codeurl}.}.

\begin{table}[h]
\centering
\small
\caption{Statistics of two real-world KGs.}
\vspace{-4mm}
\label{table:datasets}
\setlength\tabcolsep{5pt}
\def\arraystretch{1.0} % row space
\begin{tabular}{l|rrrrr}
\toprule

 & $|\mathcal{E}|$ & $|\mathcal{R}|$ & $|\mathcal{T}|$ & $\delta(q)_{avg}$ & $\delta(q)_{max}$ \\ 
 \midrule

 \textbf{FB15k237} & 14,541 & 237 & 272,115 & 37.5 & 7,614 \\
 \textbf{WN18RR} & 40,943 & 11 & 86,835 & 4.3 & 482 \\
% Dataset & \textbf{FB15k237}~\cite{bollacker2008freebase} & \textbf{WN18RR}~\cite{miller1995wordnet} \\

% \midrule
% \textit{\# of entities}  & 14,541 & 40,943 2 \\
% \textit{\# of relations} & 237 & 11  \\
% \textit{\# of triples}   & 272,115 & 86,835  \\
% \textit{Avg. degree}   & 37.5 & 4.3  \\
% \textit{Max. degree}   & 7,614 & 482  \\

\bottomrule
\end{tabular}
\end{table}

% \vspace{1mm}
\noindent
\textbf{Evaluation protocol.}
We use the protocol exactly same as that used in~\cite{sun2020reevaluation,qu2020rnnlogic}.
During the training, the accuracy of a KGC model is measured on the validation set at every epoch and the best-performing model is saved.
Then, we evaluate the best-performing model by using {\method} with varying levels of predictive sharpness and popularity-bias robustness.
We report the averaged accuracy over three runs with different random seeds.

\begin{table}[t]
% \small
\centering
\caption{The accuracy of KGC models with the varying levels of (P1) predictive sharpness and (P2) popularity-bias robustness (The bold font: best, the \underline{underline}: second-best).}
\vspace{-3mm}
\label{table:rq1-2}
\setlength\tabcolsep{7pt}
\def\arraystretch{0.9} % row space
\begin{tabular}{cc|rrrr}
\toprule

 \multicolumn{2}{c}{(P1)} & $\alpha=0.25$ & $\alpha=0.5$ & $\alpha=1$* & $\alpha=2$ \\
 \midrule

\multirow{4}{*}{\rotatebox{90}{\textbf{{FB15k237}}}} &
 RotatE & 0.544 & 0.424 & 0.314 & 0.252 \\
 & TuckEr & \textbf{0.579} & \textbf{0.463} & \textbf{0.355} & \textbf{0.294} \\
 & pLogicNet & \underline{0.549} & \underline{0.431} & \underline{0.324} & \underline{0.264} \\
 & RNNLogic & 0.534 & 0.421 & 0.320 & 0.263 \\

\midrule

\multirow{4}{*}{\rotatebox{90}{\textbf{{WN18RR}}}} &
 RotatE & \underline{0.587} & 0.520 & 0.464 & 0.431 \\
 & TuckEr & 0.560 & 0.509 & 0.469 & \underline{0.447} \\
 & pLogicNet & \textbf{0.629} & \textbf{0.546} & \underline{0.474} & 0.434 \\
 & RNNLogic & 0.575 & \underline{0.524} & \textbf{0.482} & \textbf{0.458} \\

\midrule
 \multicolumn{2}{c}{(P2)} & $\beta=0.0$* & $\beta=0.2$ & $\beta=0.4$ & $\beta=0.8$ \\
\midrule

\multirow{4}{*}{\rotatebox{90}{\textbf{{FB15k237}}}} &
 RotatE & 0.314 & 0.261 & 0.224 & 0.177 \\
 & TuckEr & \textbf{0.355} & \textbf{0.294} & \textbf{0.250} & \underline{0.194} \\
 & pLogicNet & \underline{0.324} & \underline{0.275} & \underline{0.242} & \textbf{0.204} \\
 & RNNLogic & 0.320 & 0.260 & 0.219 & 0.170 \\

\midrule

\multirow{4}{*}{\rotatebox{90}{\textbf{{WN18RR}}}} &
 RotatE & 0.464 & \underline{0.445} & \underline{0.423} & \textbf{0.375} \\
& TuckEr & 0.469 & 0.445 & 0.420 & \underline{0.367} \\
 & pLogicNet & \underline{0.474} & 0.440 & 0.409 & 0.354 \\
 & RNNLogic & \textbf{0.482} & \textbf{0.454} & \textbf{0.426} & 0.368 \\

\bottomrule
\end{tabular}
\end{table}

\subsection{Experimental Results}\label{sec:eval-result}

\vspace{1mm}
\noindent
\textbf{RQ1. Effects of predictive sharpness.}
In this experiment, we measure the accuracy of KGC models with the varying sharpness control factor $\alpha$, while fixing the level of popularity-bias robustness ($\beta=0$).
Table~\ref{table:rq1-2} shows that the accuracy of both models decreases as $\alpha$ increases (i.e., higher predictive sharpness) across all datasets.
Notably, the relative rankings of KGC models are changed.
This implies that the accuracy of KGC models is \textit{sensitive} to the level of predictive sharpness.
However, existing metrics do not reflect this perspective,
which can lead to the overestimation or underestimation of certain KGC models.
For example, RNNLogic is ranked 1st on the WN18RR when evaluated by an existing metric (MRR: $\alpha=1$).
However, its rank drops to 4th (the worst) under the evaluation with lower level of predictive sharpness ($\alpha=0.25$), which implies that RNNLogic is \textit{overestimated} by the existing metric.
Conversely, RotatE is ranked 4th on the WN18RR data under the existing metric ($\alpha=1$).
However, its rank improves to 2nd when evaluated with a lower level of predictive sharpness ($\alpha=0.25$).

% Notably, the existing metric, MRR (i.e., $\alpha=1$), 
% indicating that it strictly assess KGC models under a high predictive sharpness setting.

% \begin{table}[t]
% \small
% \centering
% \caption{Popularity bias.}
% % \vspace{-2mm}
% \label{table:rq2}
% \setlength\tabcolsep{5pt}
% \def\arraystretch{0.95} % row space
% \begin{tabular}{cc|rrrr}
% \toprule

%  Datasets & Models & $\beta=0$ & $\beta=0.2$ & $\beta=0.4$ & $\beta=0.8$ \\
%  \midrule

% \multirow{4}{*}{\rotatebox{90}{\textbf{{FB15k237}}}} &
%  RotatE & 0.314 & 0.261 & 0.224 & 0.177 \\
%  & TuckEr & \textbf{0.355} & \textbf{0.294} & \textbf{0.250} & \underline{0.194} \\
%  & pLogicNet & \underline{0.324} & \underline{0.275} & \underline{0.242} & \textbf{0.204} \\
%  & RNNLogic & 0.320 & 0.260 & 0.219 & 0.170 \\

% \midrule

% \multirow{4}{*}{\rotatebox{90}{\textbf{{WN18RR}}}} &
%  RotatE & 0.464 & \underline{0.445} & \underline{0.423} & 0.375 \\
% & TuckEr & 0.469 & 0.445 & 0.420 & \underline{0.367} \\
%  & pLogicNet & \underline{0.474} & 0.440 & 0.409 & 0.354 \\
%  & RNNLogic & \textbf{0.482} & \textbf{0.454} & \textbf{0.426} & \textbf{0.368} \\

% \bottomrule
% \end{tabular}
% \end{table}

\vspace{-2mm}
\noindent
\textbf{EQ2. Effects of popularity-bias robustness}
In this experiment, we evaluate each KGC model with varying levels of popularity-bias robustness $\beta$, while fixing the predictive sharpness level ($\alpha=1$).
As $\beta$ increases, more weights are penalized from triples involving high-popularity entities, 
i.e., assigning greater weights to the predictions on low-popularity entities.
Table~\ref{table:rq1-2} shows that the accuracy of KGC models drops significantly as $\beta$ increases, and their relative rankings change accordingly.
For example, 
on the WN18RR dataset, pLogicNet is ranked 2nd when evaluated using the existing metric ($\beta=0.0$),
but its rank drops to 4th under a more popularity-robust evaluation setting ($\beta=0.8$),
indicating that pLogicNet may be \textit{overestimated} due to reliance on popular entities.
In contrast, RotatE, ranking 4th under ($\beta=0.0$), improves to 1st when evaluated with higher popularity-bias robustness, 
suggesting that its strong performance on low-popularity entities was previously undervalued.

% These observations support our second key finding \textbf{(F2)}:
% KGC models that are highly evaluated by existing metrics tend to exhibit strong popularity bias.
% Such metrics often overestimate KGC models that perform well only on high-popularity entities,
% which can lead to overestimation or underestimation depending on the distribution of entity popularity in the dataset.
% This effect is more significant on datasets with stronger popularity bias.

\begin{figure}[t]
\begin{tikzpicture}
\begin{customlegend}[legend columns=4,legend style={align=center,draw=none,column sep=0.5ex},
        legend entries={RotatE, TuckEr, pLogicNet, RNNLogic}]
        \addlegendimage{draw=rotate,mark=o,fill=rotate, solid, line width=1pt}
        \addlegendimage{draw=tucker,fill=tucker,mark=triangle*, rotate=180, solid, line width=1pt}
        \addlegendimage{draw=plogicnet,fill=plogicnet,mark=diamond, solid, line width=1pt}
        \addlegendimage{draw=rnnlogic,fill=rnnlogic,mark=pentagon, solid, line width=1pt}
    \end{customlegend}
\end{tikzpicture}
\begin{tabular}{cc}
    \includegraphics[width=0.415\linewidth]{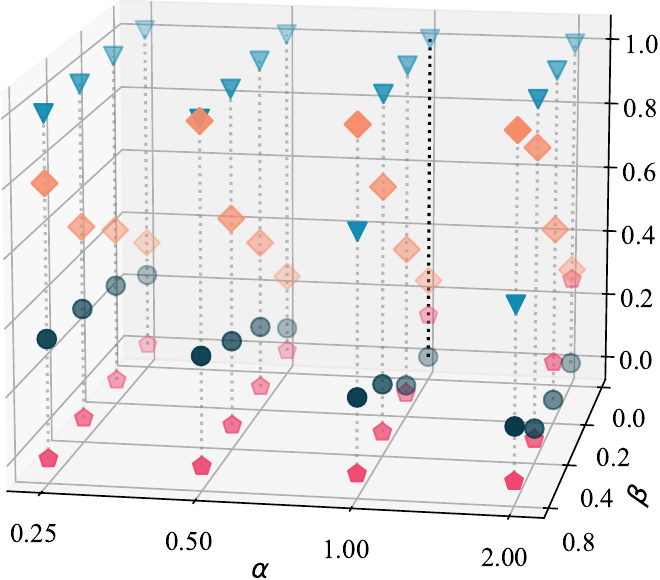} &
    \includegraphics[width=0.415\linewidth]{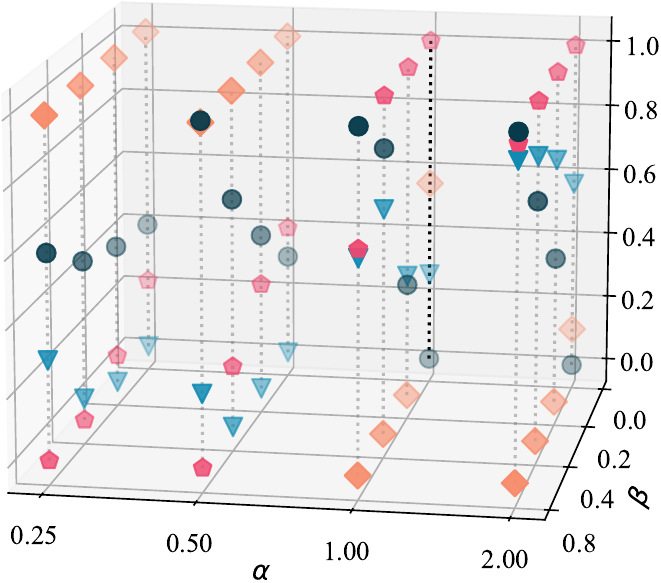} \\
    (a) FB15k237 & (b) WN18RR \\
\end{tabular}
\vspace{-3mm}
\caption{The 3-D visualized accuracy of KGC models: Different KGC models exhibit varying performance patterns depending on different evaluation perspectives.}
\vspace{-5mm}
\label{fig:eq3}
\end{figure}

\vspace{1mm}
\noindent
\textbf{EQ3. Comprehensiveness of {\method}.}
In this experiment, we evaluate how {\method} enables comprehensive evaluation of KGC models with varying levels of predictive sharpness and popularity-bias robustness.
Unlike existing metrics that consider only fixed evaluation perspectives,
{\method} allows flexible control over both dimensions, 
making researchers and practitioners easy to evaluate their KGC models under diverse real-world requirements.
Figure~\ref{fig:eq3} shows the 3-D visualized accuracy of KGC models with respect to varying levels of predictive sharpness $\alpha$ and popularity-bias robustness $\beta$.
Clearly, different KGC models exhibit varying performance patterns depending on different evaluation perspectives.
In other words, 
a model performing well under one perspective (e.g., high predictive sharpness but low popularity-bias robustness) may be outperformed by another model under a different perspective (e.g., low predictive sharpness with strong robustness to popularity bias).
For example, RNNLogic and RotatE on FB15K237
This highlights the importance of viewing the KGC model accuracy not from a single fixed perspective, but from varying perspectives.
Therefore, beyond fixed evaluation perspectives, 
{\method} provides a reliable and comprehensive understanding of KGC models.

Finally, based on the experimental results, we provide guidelines for determining appropriate levels of predictive sharpness and popularity-bias robustness.
In particular, when high predictive sharpness is desired (e.g., medical KG), $\alpha>1$ is recommended,
and when high popularity-bias robustness is preferred (e.g., facts for low-popularity entities are needed), $\beta>0.4$ is recommended.

% \begin{figure}[t]
% \begin{tabular}{cc}
%     \includegraphics[width=0.45\linewidth]{figures/FB15k237_top69.727__pLogicNet_Burlington.pdf} &
%     \includegraphics[width=0.45\linewidth]{figures/FB15k237_top30.8163__TuckER_Monster.pdf} \\
%     (a) FB15k237 & (b) WN18RR
% \end{tabular}
% \caption{Case study. }
% \label{fig:eq2-case}
% \end{figure}

% \textbf{(F3)} \textit{{\method} enables comprehensive evaluation of KGC models across varying levels of predictive sharpness and popularity-bias robustness}, where the optimal model may differ depending on the evaluation perspective.

\vspace{-1mm}
\section{Related Work}\label{sec-related} % 2.5
A handful of works have studied the KGC evaluation.
\citet{sun2020reevaluation} analyzed that a tie-breaking protocol in existing metrics can lead to overestimation of KGC model accuracy.
\citet{yang2022rethinking} observed that existing metrics provide inconsistent evaluation results in open-world settings, 
and suggested alternative metrics to alleviate the issue such as log-MRR and p-MRR.
The log-MRR and p-MRR just adopt a fixed lower level of predictive sharpness and cannot account for popularity bias.
\citet{mohamed2020popularity} proposed stratified metrics such as strat-MRR and strat-Hits@K to consider popularity bias in evaluation, 
which assign different weights to predictions on test triples based on the popularity of entities and relations.
However, the strat-MRR does not consider predictive sharpness.
\citet{hoyt2022unified} suggested an cross-dataset evaluation metric such as adjusted MRR and ZMRR to compare model accuracy not only within the same dataset but also across datasets.

\vspace{-1mm}
\section{Conclusion}\label{sec-con}
In this paper, we observe that existing rank-based metrics overlook two key perspectives for KGC evaluation:
(\textbf{A1}) \textit{predictive sharpness} and (\textbf{A2}) \textit{robustness to popularity bias}.
To reflect both aspects, we propose \textbf{{\method}}, 
a novel evaluation framework that consists of a rank transformer (RT) estimating the score of an individual prediction based on the varying levels of predictive sharpness, and a rank aggregator (RA) aggregating all the scores in a popularity-aware manner.
Experiments on two real-world KGs demonstrate that {\method} provides more reliable evaluation results depending on the required perspectives.
In future work, we plan to explore how well {\method} generalizes to other real-world KGs and KG models. 
% Moreover, we aim to extend our metric differentiable, enabling KGC models to automatically learn preferred perspectives and thereby aligning evaluation with practical needs.

% \section{GenAI Usage Disclosur}\label{sec-genai}
% In accordance with the ACM authorship policy, we disclose the usage of generative AI tools (e.g., ChatGPT) as follows.
% \begin{itemize}[leftmargin=10pt]
%     \item \textbf{GenAI usage in writing}: ChatGPT was only used to review grammatical consistency during the writing the manuscript.
%     \item \textbf{GenAI usage in data processing}: ChatGPT was only used for generating code for drawing figures (e.g., matplotlib.pyplot).
%     \item \textbf{Author responsibility}: All uses of GenAI were limited to assistant roles. We conducted final verification and refinement of all GenAI generated results, analyses, and textual content.
% \end{itemize}
\vspace{-1mm}
\begin{acks}
This work was supported by the National Research Foundation of Korea (NRF) grant funded by the Korea government (MSIT) (RS-2024-00459301).
\end{acks}


\bibliographystyle{ACM-Reference-Format}
\bibliography{bibliography.bib}

% \appendix

\end{document}